# DisorderUnetLM: Validating ProteinUnet for efficient protein intrinsic disorder prediction

Krzysztof Kotowski[1], Irena Roterman[2] and Katarzyna Stapor[1,*]

[1]Department of Applied Informatics, Silesian University of Technology, Akademicka 16, 44-100 Gliwice, Poland

[2]Department of Bioinformatics and Telemedicine, Jagiellonian University Medical College, Medyczna 7, 30-688, Kraków, Poland

*Corresponding author.

## ABSTRACT

The prediction of intrinsic disorder regions has significant implications for understanding protein functions and dynamics. It can help to discover novel protein-protein interactions essential for designing new drugs and enzymes. Recently, a new generation of predictors based on protein language models (pLMs) is emerging. These algorithms reach state-of-the-art accuracy without calculating time-consuming multiple sequence alignments (MSAs). The article introduces the new DisorderUnetLM disorder predictor, which builds upon the idea of ProteinUnet. It uses the Attention U-Net convolutional neural network and incorporates features from the ProtTrans pLM. DisorderUnetLM achieves top results in the direct comparison with recent predictors exploiting MSAs and pLMs. Moreover, among 43 predictors from the latest CAID-2 benchmark, it ranks 1st for the Disorder-NOX subset (ROC-AUC of 0.844) and 10th for the Disorder-PDB subset (ROC-AUC of 0.924). The code and model are publicly available and fully reproducible at doi.org/10.24433/CO.7350682.v1.

## Introduction

Functional regions in proteins can either be structured or disordered, and these can be considered as two fundamental classes of functional building blocks of proteins [1]. Protein intrinsic disordered regions are segments of proteins that have ambiguous three-dimensional structures in isolated conditions [2], [3]. They are important in identifying functions of a protein, because, due to their high flexibility, they can engage in numerous different chemical interactions, such as regulation, signalling, transcriptional, and translational processes [4]. Disordered regions can be resolved experimentally, e.g., using nuclear magnetic resonance (NMR) spectroscopy, but it is time-consuming and expensive [3]. Thus, the prediction of disordered regions from their amino acid sequences has become a popular research area in bioinformatics and benchmarks like CAID (Critical Assessment of Intrinsic Protein Disorder) [5], [6] have emerged to assess and compare different predictors. Accurate prediction of disorder regions can help to discover novel functions or protein-protein interactions essential to designing new drugs, therapies, or enzymes.

The simplest predictors are based on the idea that disordered regions usually contain a significantly larger proportion of small and hydrophilic amino acids and proline residues than structured regions [4]. There are also classic approaches based on typical patterns of neighbouring amino acids, i.e., n-grams (or k-mers) [7]. However, machine learning and deep learning models using evolutionary information, e.g., PSSM (position-specific scoring matrices) [8] or HHblits (iterative protein sequence search according to the hidden profile) [9] have quickly dominated the benchmarks [6], [10]–[14]. These approaches can learn complex patterns from similar sequences and capture subtle features of intrinsic disorder regions. Evolutionary information provides much better features than amino acid sequences alone [14] but is very computationally expensive to obtain. Recently, pre-trained language models based on the idea of attention and transformers [15] have been adopted for the protein secondary structure and disorder prediction and they show state-of-the-art results [16]–[19]. They are called protein language models and they implicitly embed the evolutionary information in their compact feature space, which allows them to provide better features in a fraction of the time needed for classic multiple sequence alignments.

The current article follows this latest trend and presents DisorderUnetLM – a convolutional Attention U-Net [20] architecture using features from the ProtTrans protein language model [17]. The idea is based on our previous ProteinUnetLM network [18] which showed state-of-the-art results in protein secondary structure prediction. Main novelties are related to (1) the adapted output of the network, i.e., binary disorder prediction instead of 8-class secondary structure prediction; (2) additional mechanisms to prevent overfitting for smaller datasets, i.e., fewer convolutional units, stronger dropout, and earlier stopping; (3) the ensembling procedure adopted from the first version of ProteinUnet [21] to boost performance in the CAID-2 benchmark [6] and in the latest CAID-3 challenge (caid.idpcentral.org/challenge), for which DisorderUnetLM has been submitted. The proposed method is thoroughly validated and compared with the state-of-the-art using procedures introduced by flDPnn [10], SETH [16], and especially, the CAID-2 benchmark [6].

## Methods

DisorderUnetLM was implemented in the environment containing Python 3.8 with TensorFlow 2.9 accelerated by CUDA 11.7 and cuDNN 8. The inference code and trained models are available on the CodeOcean platform ([doi.org/10.24433/CO.7350682.v1](doi.org/10.24433/CO.7350682.v1)) ensuring high reproducibility of the results.

### Datasets

There are 6 datasets used in our study. They are listed in Table 1 with source links, percentages of disordered residues, and numbers of training, validation, and test sequences as defined by the datasets' authors. There are 3 training sets, namely flDPnn [10], CheZOD [16], and IDP-CRF [22], and they are used to train 3 different versions of the DisorderUnetLM model for a fair and direct comparison with the corresponding predictors (i.e., flDPnn, SETH, and IDP-CRF) in the Results section. The final DisorderUnetLM version submitted to the CAID-3 challenge uses all 8799 sequences for training.

The smallest training set of 445 sequences belongs to the flDPnn dataset. It was introduced together with the predictor of the same name [10]. It is the only dataset that explicitly defines a validation set (of 100 sequences). The test set of 176 sequences has <25% similarity with the training set according to the CD-HIT algorithm [23]. Together, this gives a compact benchmarking dataset with 721 sequences from the DisProt 7.0 database [24] with nearly 25% disordered residues.

The CheZOD (Chemical shift Z-score for quantitative protein Order and Disorder assessment [25]) dataset with 1174 training and 117 testing sequences is taken from the article about the SETH predictor [16]. The training set is constructed such that all proteins have less than 20% pairwise sequence similarity with the testing set according to the MMSeqs2 high-sensitivity search [26]. Unlike in other datasets, the CheZOD ground truth is not binary. It quantifies the degree of disorder based on the assigned Z-scored nuclear magnetic resonance chemical shifts, ranging between -4 for complete disorder and 16 for complete order, with a value of 8 corresponding to an intermediate position in the disorder continuum. For purposes of this study, the CheZOD ground truth was binarized, so all residues with CheZOD scores higher than 8 are marked as ordered and the rest are marked as disordered.

The IDP-CRF is the largest training set in the study with 4590 sequences from MobiDB [27] and 683 sequences from DisProt 7.0 [24] database. Sequences in this dataset have less than 25% similarity between each other according to the Blastclust algorithm [28]. A subset of 494 non-redundant sequences from the MxD dataset [29] of 514 sequences (319 from DisProt 5.0 [30] and 205 from Protein Data Bank [31]) was used by the authors of IDP-CRF as an independent test set (reportedly, after removing redundancy with the IDP-CRF using Blastclust with similarity of 25%). However, our verification using MMseqs2 with the same level of similarity showed that only less than 100 sequences from the MxD dataset showed no similarity to the IDP-CRF dataset.

The testing datasets from CAID [5] and CAID-2 [6] benchmarks contain 652 and 348 sequences from the DisProt database, respectively. Specifically, these numbers concern the *Disorder-PDB* versions of the benchmark which only include ordered regions if they are observed in the Protein Data Bank database [31]. There are also subsets of sequences where all residues are marked as structured unless they were experimentally annotated as disordered (called *Disorder* in CAID and *Disorder-NOX* in CAID-2). Results for both versions are reported in our study.

*Table 1. List of datasets used in the study with numbers of sequences in training, validation, and testing sets as defined by their authors. The values in parentheses mark the ratio of disordered residues in the set.*

| Dataset name and reference | Link to download | Number of sequences | | | |
|---|---|---|---|---|---|
| | | Training | Validation | Testing | Total |
| flDPnn [10] | [biomine.cs.vcu.edu/servers/flDPnn/](biomine.cs.vcu.edu/servers/flDPnn/) | 445 (22.91%) | 100 (32.61%) | 176 (26.86%) | 721 (24.66%) |
| CheZOD (binarized) [16] | [github.com/DagmarIlz/SETH](github.com/DagmarIlz/SETH) | 1174 (26.18%) | - | 117 (68.77%) | 1291 (29.83%) |
| IDP-CRF [22] | [mdpi.com/1422-0067/19/9/2483/s1](mdpi.com/1422-0067/19/9/2483/s1) | 5273 (9.60%) | - | - | 5273 (9.60%) |
| MxD [29] | [biomine.cs.vcu.edu/servers/MFDp/MxD.txt](biomine.cs.vcu.edu/servers/MFDp/MxD.txt) | - | - | 514 (21.83%) | 514 (21.83%) |
| CAID Disorder [5] | [doi.org/10.24433/CO.3610625.v1](doi.org/10.24433/CO.3610625.v1) | - | - | 652 (16.23%) | 652 (16.23%) |
| CAID-2 Disorder [6] | [caid.idpcentral.org/assets/sections/challenge/static/references/2/disorder_pdb.fasta](caid.idpcentral.org/assets/sections/challenge/static/references/2/disorder_pdb.fasta) | - | - | 348 (12.92%) | 348 (12.92%) |
| Total | | 6892 (12.55%) | 100 (32.61%) | 1807 (18.09%) | 8799 (14.66%) |

**Attention U-Net for protein intrinsic disorder prediction**

U-Net is a state-of-the-art architecture in image segmentation tasks [32]–[34] and we previously successfully introduced it into the domain of protein secondary structure prediction by creating the ProteinUnet model [14], [21]. For the disorder prediction, we base on our latest Attention U-Net architecture of ProteinUnetLM [18] using features from the ProtTransT5-XL-U50 protein language model [17] as input. The selection of this specific protein language model (pLM) is dictated by conclusions from our previous study on protein secondary structure prediction, where ProtTrans showed better results, faster processing, shorter feature vectors, and higher sequence length limits than ESM-1b [35]. Additional literature review on the latest advancements in this topic, especially the second version of the ESM model [36], revealed mixed results in comparison to the ProtTrans. Some articles still favor ProtTrans over ESM-2 [37] while others were less conclusive [38], [39]. Finally, we keep using ProtTrans following the claim proposed in the latest paper by Rost's lab that "choice of pLM is relevant but not critical" [40].

The detailed architecture of DisorderUnetLM is presented in Figure 1. The only input of the model is a set of features from the ProtTrans pLM. Unlike in ProteinUnetLM, we do not use the additional input with one-hot encoded amino acid sequences, because the ablation study in Supplementary Table S1 showed no advantage of such an approach. Moreover, we decreased the number of convolutional layers at each level or U-Net from 64/128 to 32/64 and increased the dropout rate from 0.1 to 0.25 to avoid overfitting due to smaller training sets and smaller output dimensionality (8-class secondary structure vs binary disorder states). All other hyperparameters are the same as in the ProteinUnetLM. Specifically, we have 2 convolutions with 1D kernels of length 7 and ReLU activations in all blocks. Overall, a single DisorderUnetLM model has 628,710 trainable parameters.

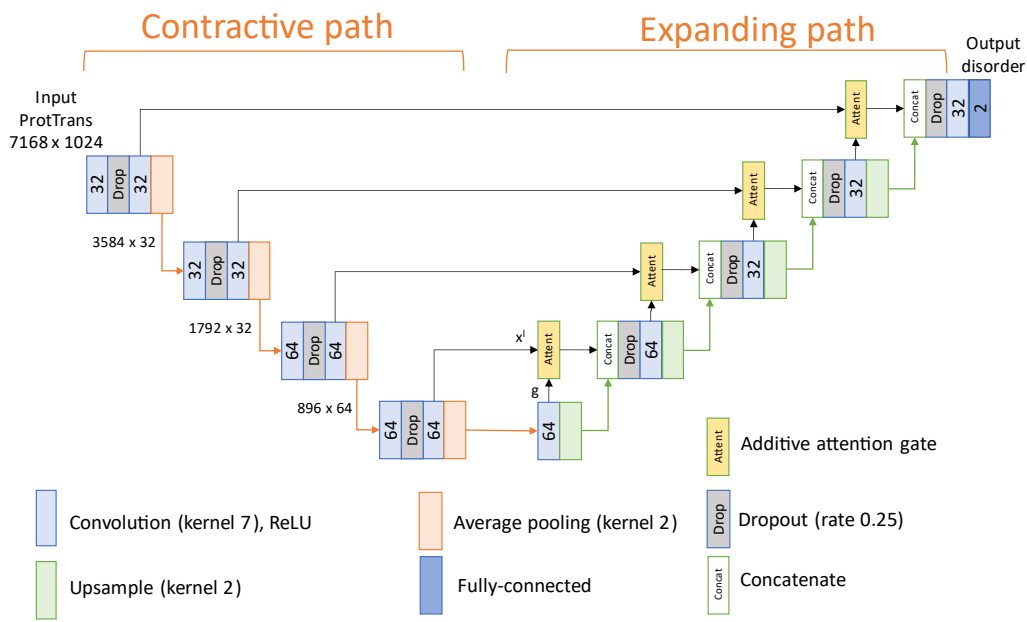

**Figure 1**. *The detailed architecture of DisorderUnetLM. Symbols $x^l$ and g correspond to the input features and attention coefficients as denoted in Figure 2.*

The network learns higher-level features in convolutional contractive paths, concatenates them, and passes them to the additive attention gates (AGs) presented in Figure 2. AGs learn to select and focus (give attention) on the most important features passed by skip connections [15], [20]. The output of the AG can be treated as a saliency map which gives high weights to relevant features and low weights to irrelevant ones. Information extracted from lower-scale features is used as a gating signal to disambiguate irrelevant and noisy responses in skip connections. AGs are active both during backward pass (training) and forward pass (prediction), and their role is to filter irrelevant parts of the input features. This should allow for better generalization of the network and improved robustness to noisy data. Finally, the filtered features are passed to the convolutional expanding path that learns to predict the disorder probability as the output layer with softmax activation connected to the last up-block (Figure 1).

DisorderUnetLM takes a sequence of feature vectors $X = (x_1, x_2, x_3, ..., x_N)$ as input, where $x_i$ is the feature vector corresponding to the *i*th residue, and returns a vector $Y = (y_1, y_2, y_3, ..., y_N)$ as output, where $y_i$ is a probability of *i*th residue being in the disordered state. If this probability is greater than 0.5 the residue is marked as disordered. The input sequence length is limited to 7168 which covers all proteins used in this study and nearly all proteins available in the latest DisProt [41] database (excluding only Titin with 34350 amino acids). For each amino acid, there are 1024 features from the ProtTransT5-XL-U50 protein language model [17]. Each feature is standardized across all training residues to ensure a mean of 0 and a standard deviation of 1. As a rule of thumb, standardization of the input data is usually beneficial for the process of learning, as proven in seminal works by LeCun and Glorot [42], [43].

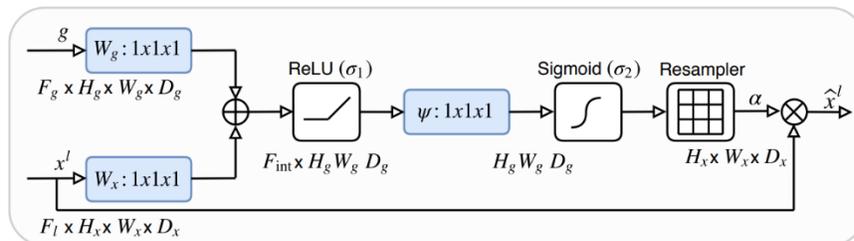

***Figure 2***. *Schematic of the additive attention gate (AG) . Input features ($x^l$) are scaled with attention coefficients (α) computed in AG. Spatial regions are selected by analysing both the activations and contextual information provided by the gating signal (g) which is collected from a coarser scale. Grid resampling of attention coefficients is done using trilinear interpolation. Source: [20]*

**Training procedures and loss function**

Following the ProteinUnetLM [18] procedures, DisorderUnetLM was trained to simultaneously minimize the binary cross-entropy (BCE, Equation 1) and maximize the Matthews correlation coefficient (MCC, adapted to the continuous probabilities as defined in Equation 2) by defining the loss function as a difference between average BCE and average MCC across a training batch (Equation 3). The analysis for ProteinUnetLM showed that adding the MCC term had a strong effect on the performance for very rare secondary structures (far less than 5% of all residues). In the disorder prediction task, the disorder ratios are much higher (between 10% and 30%), so this effect may be less pronounced. Nonetheless, we keep the MCC term for consistency.

$$BCE = \boldsymbol{y}\log(\hat{\boldsymbol{y}}) + (1-\boldsymbol{y})\log(\hat{\boldsymbol{y}}), \quad \text{where } \boldsymbol{y} \text{ is a target vector and } \hat{\boldsymbol{y}} \text{ is a model output,} \quad (1)$$

$$MCC = \frac{TP \times TN - FP \times FN}{\sqrt{(TP+FP)(TP+FN)(TN+FP)(TN+FN)} + e} \quad (2)$$

where $TP = \boldsymbol{y} \cdot \hat{\boldsymbol{y}}, TN = (\boldsymbol{1}-\boldsymbol{y}) \cdot (\boldsymbol{1}-\hat{\boldsymbol{y}}), FP = (\boldsymbol{1}-\boldsymbol{y}) \cdot \hat{\boldsymbol{y}}, FN = \boldsymbol{y} \cdot (\boldsymbol{1}-\hat{\boldsymbol{y}})$, and $e$ is a very small number preventing division by zero,

$$Loss = BCE - MCC \quad (3)$$

Adam optimizer [44] is used with a batch size of 8 and an initial learning rate of 0.001. The learning rate is reduced by a factor of 10 when there is no improvement in the validation loss in consecutive epochs. The training is stopped when the validation loss is not improving for 5 epochs and the checkpoint with the lowest validation loss among all epochs is selected as the final model. Such early stopping was used to avoid any potential overfitting for smaller datasets.

*Ensembling for the CAID-2 benchmark and the CAID-3 challenge*

To train the final DisorderUnetLM model for purposes of the CAID-3 challenge, we use the ensembling procedure introduced in ProteinUnet [21]. All collected datasets (including test sets) are merged into a single training set of 8799 sequences and a 10-fold stratified cross-validation is performed. The folds are stratified based on the sequence lengths and ratios of disordered residues. The 10 resulting models (each trained on 9 folds and validated on the remaining one) are ensembled by taking the average of their output probabilities for each residue. If the average probability is greater than 0.5 the residue is marked as disordered. Note that this final ensemble is not tested in the current article as all available data are used for training to maximize the result in the upcoming CAID-3 challenge. The same ensembling procedure was used to evaluate the algorithm on the CAID-2 test set (excluding this test set from the training data). In this case, the training set had 8451 sequences.

**Evaluation procedures**

We directly follow the evaluation procedures introduced in referenced papers (i.e., flDPnn [10] and SETH [16]) and in the CAID-2 competition [6], i.e., we use the same metrics and training\test subsets (as defined in Table 1.). The popular area under receiver operating characteristic (ROC-AUC) is used as the main metric in all of those procedures. This metric operates directly on the output probabilities and, unlike MCC and F1-scores used in the flDPnn evaluation, it is independent of the additional thresholding method (i.e., the simple threshold of 0.5). ROC-AUC is not an ideal metric and gives only a partial insight into performance [45], but since we target the CAID-3 competition with DisorderUnetLM, we focus on ROC-AUC as the primary metric used in this challenge.

# Results

In this section, DisorderUnetLM is benchmarked against evaluation procedures proposed by authors of flDPnn [10] and SETH [16] predictors, and against more than 40 predictors from the latest CAID-2 competition [6]. Detailed numerical values for all visual comparisons presented in the main text are given in Supplementary Tables S2-S5.

### Validation using flDPnn procedures

Following the procedures from the article about the flDPnn predictor, a single DisorderUnetLM trained on the flDPnn training set was compared with 5 predictors (flDPnn [10], ESpritz-D [46], SPOT-Disorder-Single [47], IUPred2A-long [48], and IUPred-2A-short [48]) on the flDPnn test set (Figure 3) and with 10 predictors (flDPnn [10], flDPlr [10], RawMSA [49], ESpritz-D [46], DisoMine [50], SPOT-Disorder2 [11], AUCpreD [51], SPOT-Disorder-Single [47], AUCpreD-np [51], PreDisorder [52]) on the well-established CAID Disorder-PDB test set (Figure 4). The existing redundancy between the CAID test set and the flDPnn training set was not removed to ensure the exact same validation procedure and to allow for a direct comparison with results reported in the flDPnn paper (released before the CAID sequences were published). DisorderUnetLM results on the flDPnn test set are comparable to the results of the flDPnn predictor. Our model is slightly better in terms of F1-score (0.629 vs 0.626), but slightly worse in ROC-AUC (0.835 vs 0.839) and MCC (0.478 vs 0.491) metrics. However, DisorderUnetLM shows a clear advantage over the flDPnn predictor on the larger CAID test set in every metric, F1-score (0.516 vs 0.483), ROC-AUC (0.826 vs 0.814), and MCC (0.414 vs 0.370). Both DisorderUnetLM and flDPnn overcome other predictors by a large margin in all metrics.

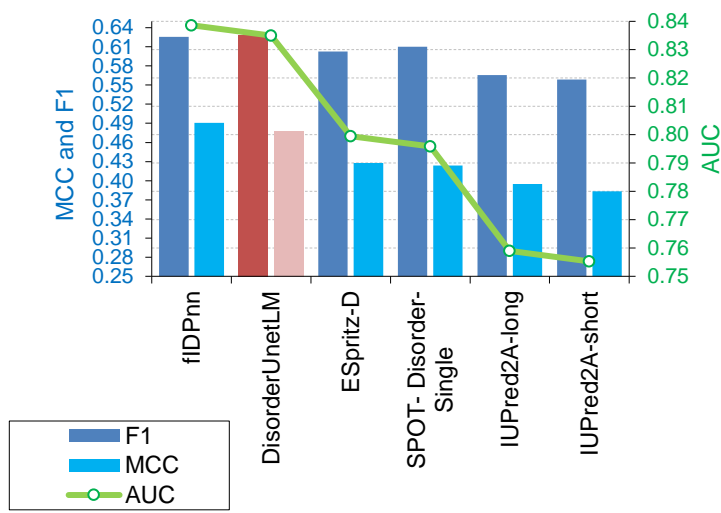

*Figure 3*. Comparison of DisorderUnetLM (marked in red) with 5 other predictors on the flDPnn test set. The visualization is adapted from the article about the flDPnn predictor [10].

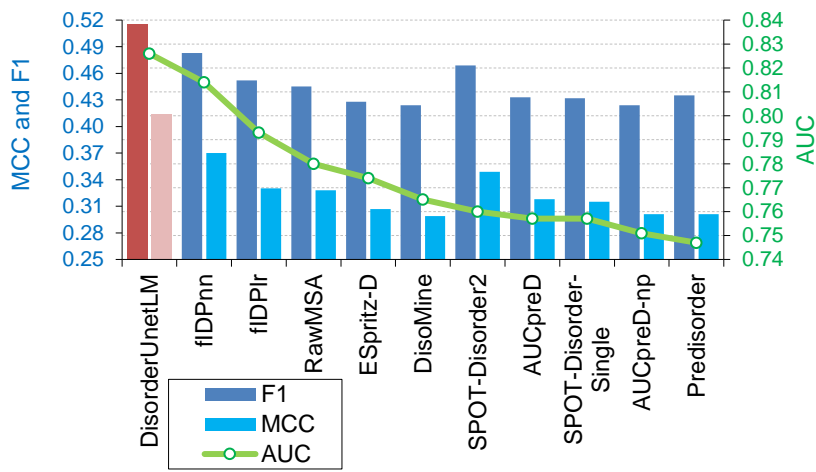

*Figure 4*. Comparison of DisorderUnetLM (marked in red) with 10 other predictors on the CAID test set. The visualization is adapted from the article about the flDPnn predictor [10].

**Validation using SETH procedures**

Following the procedures from the article about the SETH predictor, the binarized CheZOD training set was used to train a single DisorderUnetLM model using a random 10% of sequences as a validation set. In Figure 5, ROC-AUC scores on the binarized CheZOD test set were compared with 15 selected predictors described in the SETH article. Besides DisorderUnetLM, two predictors use features from protein language models – SETH and ADOPT-Esm1b. They show a clear advantage over the 9 predictors using evolutionary information from multiple sequence alignments (ODiNPred [12], SPOT-Disorder [53], AlphaFold2-rsa-25 [13], AUCpreD [51]. MetaDisorder [54], MFDp2 [55], PrDOS [56], DISOPRED3 [57], flDPnn [10]) and remaining 4 using only amino acids sequences (AUCpreD-noEvo [51], IUPred [58], DISPROT VSL2b [59]). DisorderUnetLM achieves the same ROC-AUC score (0.910) as SETH, using the same ProtTrans protein language model. However, SETH was trained using continuous CheZOD scores which should give some advantage on the CheZOD test set, thanks to more detailed information beyond the binarized disorder status. Thus, DisorderUnetLM proved to be at least as effective as SETH.

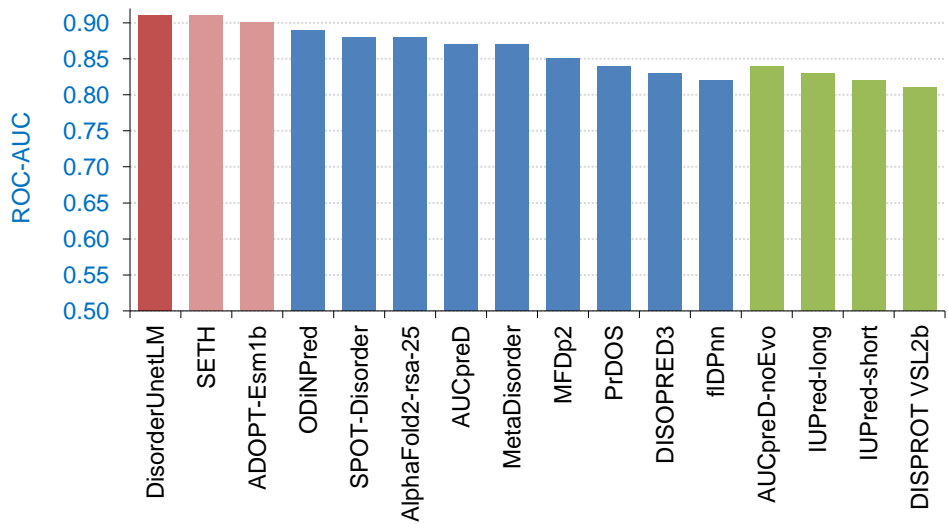

*Figure 5. Comparison of DisorderUnetLM (marked in red) with 15 other selected predictors on the binarized CheZOD test set. 2 predictors use features from protein language models (marked in light red), 9 predictors explicitly use evolutionary information (marked in blue) and 4 predictors use only classic features from amino acid sequences (marked in green).*

**Validation using CAID-2 procedures**

Finally, DisorderUnetLM was compared on the CAID-2 dataset with 41 predictors for which ROC-AUC values are available at the official CAID portal (caid.idpcentral.org/challenge/results). The list was also extended with results of the NetSurfP-3.0 web server (services.healthtech.dtu.dk/services/NetSurfP-3.0) [60] based on the ESM-1b protein language model [35].

To maximize the performance of the DisorderUnetLM model, all the collected datasets (excluding the CAID-2 test set) were used to train an ensemble of 10 classifiers as described in Methods. Additionally, performance was also verified on the non-redundant CAID-2 subsets (see Supplementary Material) created by removing 46 sequences of similarity higher than 25% to the merged training set according to the MMseqs2 clustering. To generate comparisons for the non-redundant set, the official CAID-2 predictions repository was used (available under the link: caid.idpcentral.org/assets/sections/challenge/static/predictions/2/predictions.zip). However, 11 algorithms had to be omitted in this comparison because of missing entries for some sequences.

As presented in Figure 6, the ensembled DisorderUnetLM achieved the 10th best ROC-AUC out of 43 predictors for CAID-2 Disorder-PDB (0.924 vs 0.949 for the best SPOT-Disorder2 [11]). The ordering of DisordetUnetLM among top predictors does not change after removing redundant sequences (see Supplementary Figure S1). As presented in Figure 7, for the smaller CAID-2 Disorder-NOX test set, DisorderUnetLM achieved the best ROC-AUC among all predictors (0.844 vs 0.838 for the second-best Dispredict3 [61]). Again, the ordering of top predictors does not change after removing redundant sequences and DisorderUnetLM stays in the first place with just slightly lower 0.834 ROC-AUC (see Supplementary Figure S2). Disorder-NOX has a higher ratio of ordered residues (negatives) than Disorder-PDB (80.5% vs 71.7%). It is because all residues without structural annotation available in PDB (or DisProt) are marked as ordered in Disorder-NOX while they are ignored in Disorder-PDB. The advantage of DisorderUnetLM over other predictors for the Disorder-NOX subset suggests that our model has better precision (fewer residues marked as disordered) for those uncertain regions.

Figure 8 and Figure 9 present detailed visual comparisons of predictions for sequences DP02553 and DP02996 from the CAID-2 Disorder-NOX test set for the best 3 models, i.e., DisorderUnetLM, flDPnn, and Dispredict3. The most prominent observation is that predictions from DisorderUnetLM are smoother, without sudden changes of disorder probability between consecutive residues. Its predictions for DP02553 are relatively close to the other models, but for DP02996 they are drastically different with much higher disorder probabilities. This qualitative comparison shows that DisorderUnetLM is not just an incremental improvement over other predictors but may offer a very different view on some proteins.

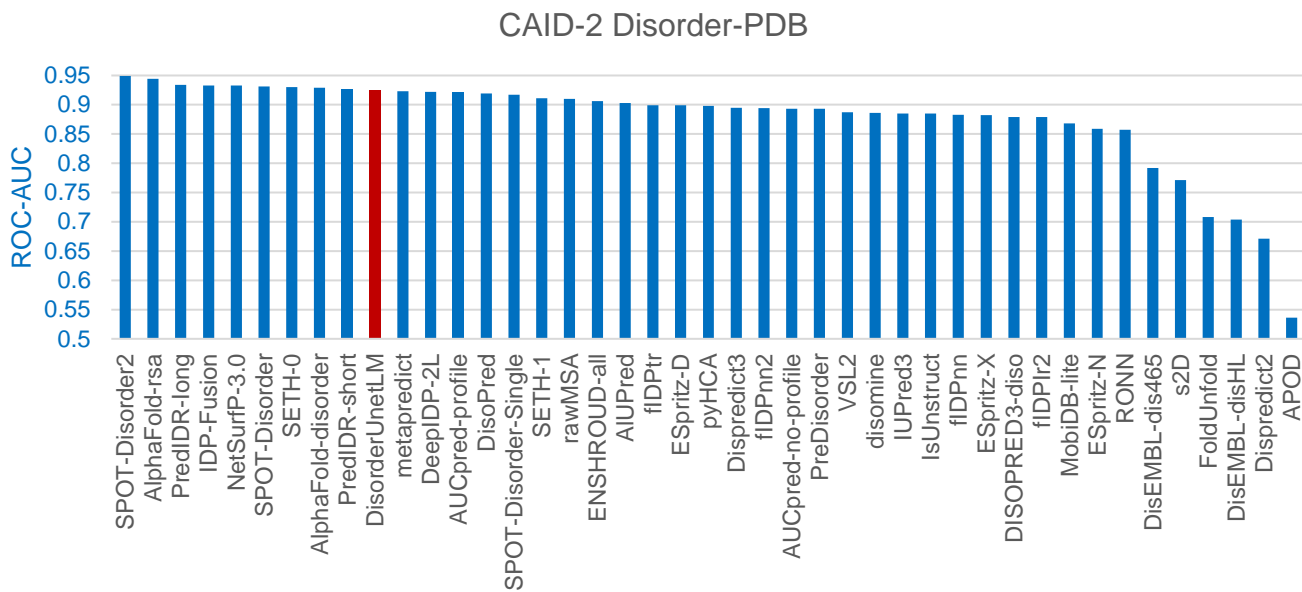

**Figure 6**. *The comparison of ROC-AUC for 43 methods on the CAID-2 Disorder-PDB dataset. Our proposed DisorderUnetLM is marked in red.*

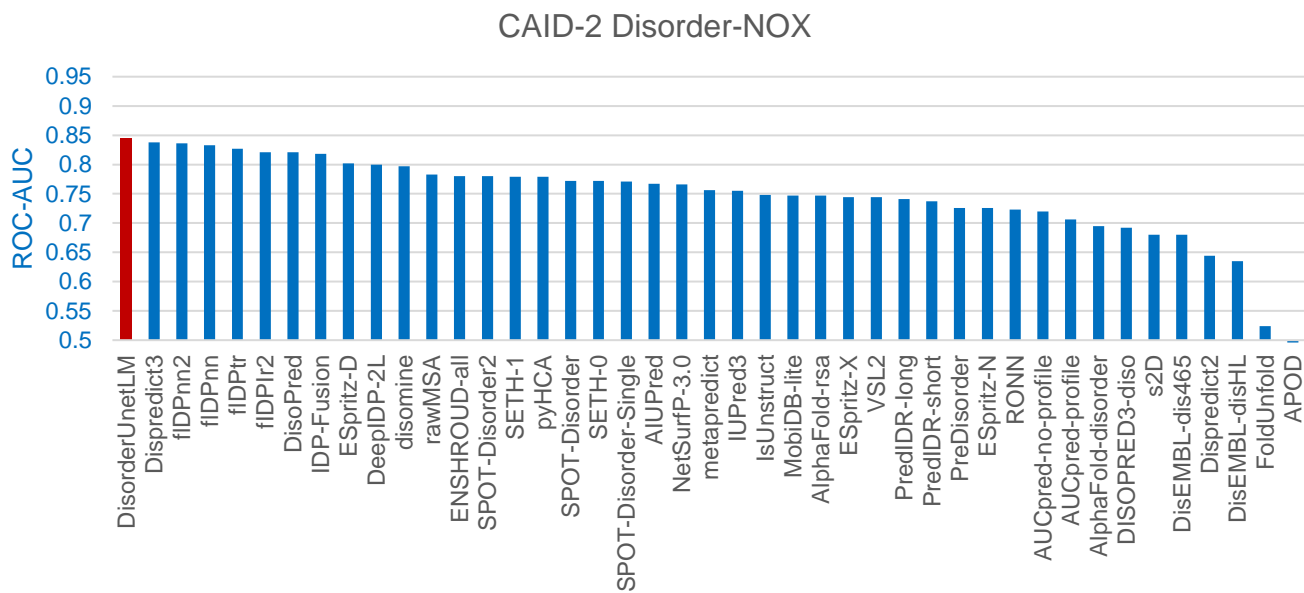

**Figure 7**. *The comparison of ROC-AUC for 43 methods on the CAID-2 Disorder-NOX dataset. Our proposed DisorderUnetLM is marked in red.*

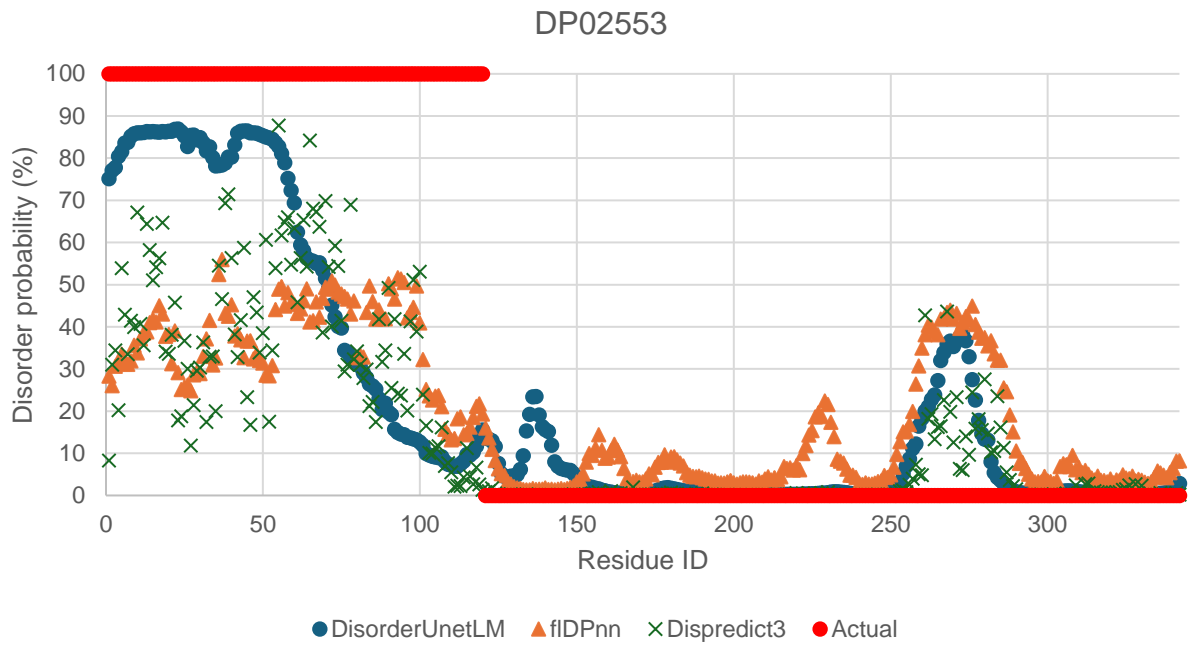

**Figure 8**. *The visual comparison of predictions for DisorderUnetLM, flDPnn, and Dispredict3 models for the DP02553 sequence from the CAID-2 Disorder-NOX test set.*

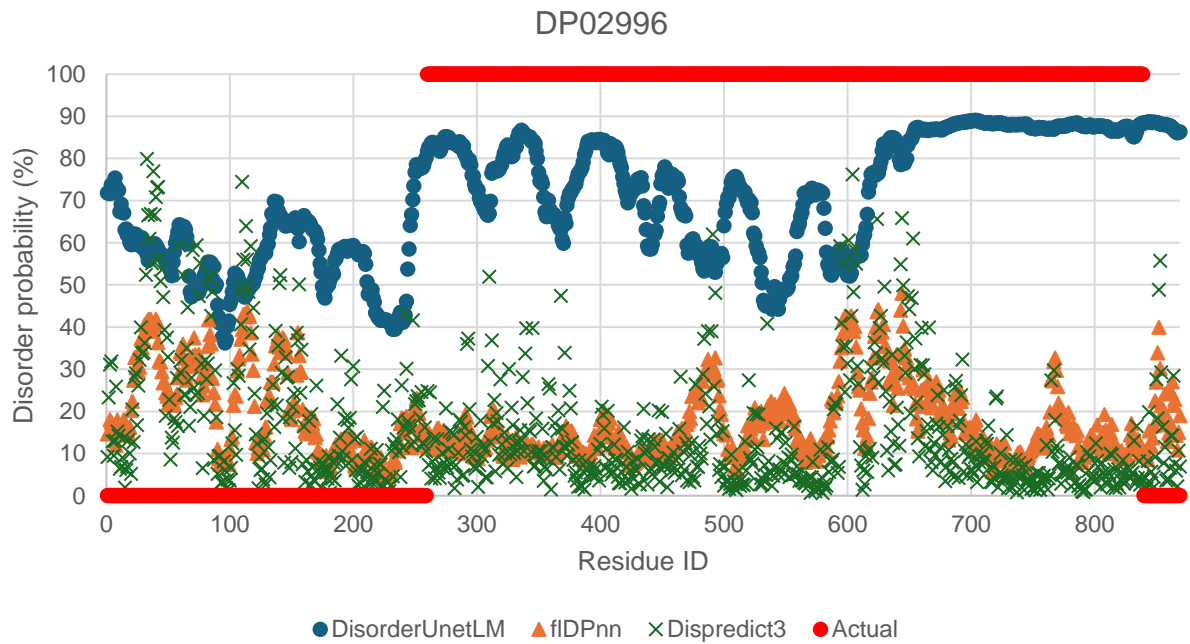

**Figure 9**. *The visual comparison of predictions for DisorderUnetLM, flDPnn, and Dispredict3 models for the DP02996 sequence from the CAID-2 Disorder-NOX test set.*

## Conclusions

The Attention U-Net using features from the ProtTrans protein language model (DisorderUnetLM) proved their high utility in the task of protein intrinsic disorder prediction, just like it recently did in the domain of protein secondary structure prediction (ProteinUnetLM) [18]. In this study, DisorderUnetLM is compared with more than 50 predictors in 5 different evaluation scenarios. It shows top results in direct comparisons with predictors using classic and evolutionary features, and with the SETH [16] predictor using features from the same ProtTrans model. Moreover, it is ranked in the top 10 best-performing methods among 43 predictors in the CAID-2 benchmark (10$^{th}$ place in Disorder-PDB with ROC-AUC of 0.924 and 1$^{st}$ place in Disorder-NOX test sets with ROC-AUC of 0.844). It was confirmed also on the non-redundant versions of those benchmarks. The advantage of DisorderUnetLM over other predictors for the Disorder-NOX subset suggests that our model has better precision (fewer residues marked as disordered) for those uncertain regions. It may be a useful feature in applications where only genuine disordered regions should be reported. Results for smaller test sets (flDPnn and CheZOD) show no substantial improvement. However, the results of DisorderUnetLM are comparable to the best predictors and the question arises if they can be even improved without using additional data or if they are already at their upper bound. In summary, DisorderUnetLM shows a potential to perform well in the upcoming CAID-3 challenge for which it was submitted.

The convolutional Attention U-Net architecture is characterized by relatively fast training and inference as compared to recurrent neural networks for protein structure prediction [18]. Additionally, DisorderUnetLM does not use computationally expensive evolutionary features but the output of the ProtTrans model - calculated in a fraction of a second per sequence. It is useful in large-scale predictions and low-grade devices. We share the complete code and models on the CodeOcean platform to support the reproducibility of our work and to encourage the community of protein scientists to use our method in their research, e.g., to study functions of proteins [62], protein-protein interactions [63], or cellular signalling and regulation [64]. In the future, the Attention U-Net architecture can be easily adapted to many other use cases like a prediction of continuous CheZOD scores [16], binding sites, or linkers.

## Data and Software Availability

The inference code and trained models are available on the CodeOcean platform ensuring high reproducibility of the results: doi.org/10.24433/CO.7350682.v1. The datasets used in the study are publicly available under the links given in Table 1.

## Supplementary material

The exact version of the ProtTransT5-XL-U50 model used in the study can be downloaded from https://huggingface.co/Rostlab/prot_t5_xl_uniref50/blob/main/pytorch_model.bin and has been run using ProtTransT5XLU50Embedder class from bio_embeddings Python library in version 0.2.2 (https://github.com/sacdallago/bio_embeddings/releases/tag/v0.2.2).

**Supplementary Table S1**. Ablation study of DisorderUnetLM trained on the flDPnn training set and tested on the CAID dataset.

| Model | MCC | F1 | ROC-AUC |
|---|---|---|---|
| DisorderUnetLM | 0.411 | 0.516 | 0.825 |
| With the additional sequence of amino acids on input | 0.385 | 0.493 | 0.817 |
| With the additional sequence of amino acids on input and 64 layers, like in ProteinUnetLM | -0.002 | 0.0 | 0.797 |

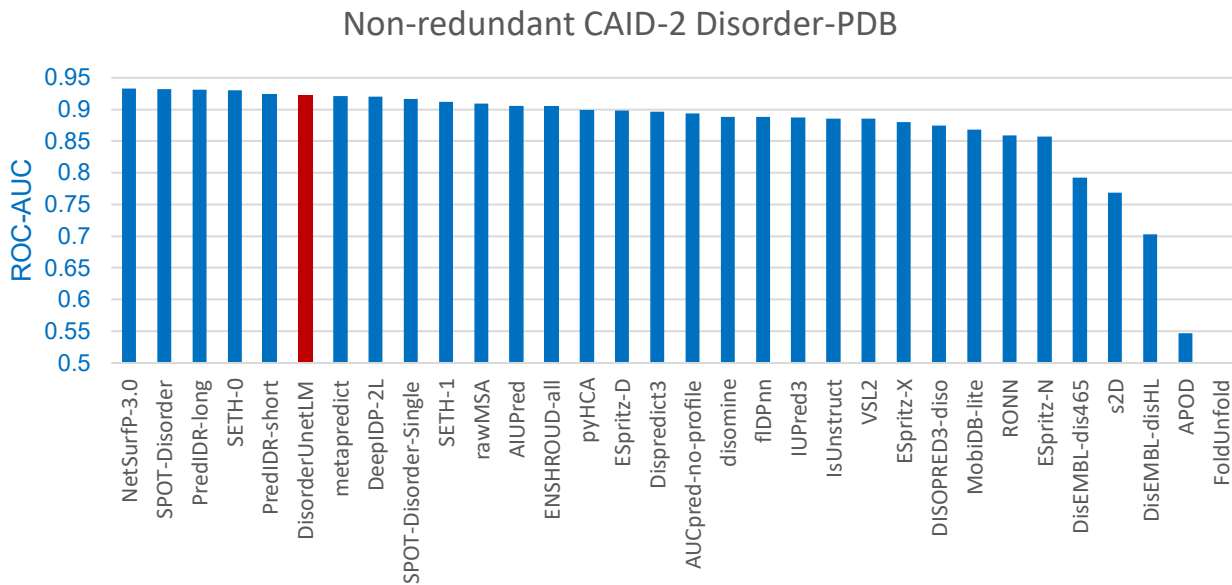

**Supplementary Figure S1**. The comparison of ROC-AUC for 32 methods on the non-redundant CAID-2 Disorder-PDB dataset. Our proposed DisorderUnetLM is marked in red.

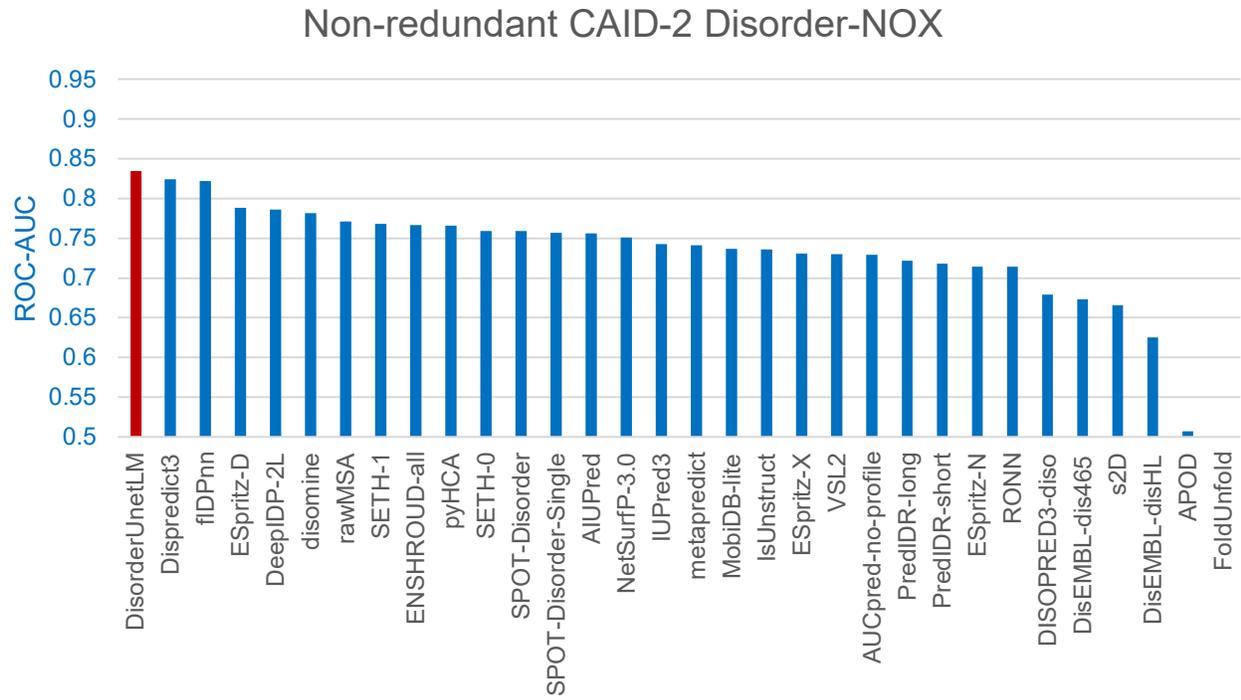

**Supplementary Figure S2**. *The comparison of ROC-AUC for 32 methods on the non-redundant CAID-2 Disorder-NOX dataset. Our proposed DisorderUnetLM is marked in red.*

**Supplementary Table S2**. Detailed ROC-AUC, F1, and MCC metrics values for the evaluation on the flDPnn test set. The best values for each metric are boldfaced. The proposed DisorderUnetLM is underlined.

| Method | ROC-AUC | F1 | MCC |
| --- | --- | --- | --- |
| fIDPnn | **0.839** | 0.626 | **0.491** |
| DisorderUnetLM | 0.835 | **0.629** | 0.478 |
| ESpritz-D | 0.799 | 0.603 | 0.428 |
| SPOT- Disorder-Single | 0.796 | 0.610 | 0.424 |
| IUPred2A-long | 0.759 | 0.566 | 0.395 |
| IUPred2A-short | 0.755 | 0.559 | 0.383 |

**Supplementary Table S3**. Detailed ROC-AUC, F1, and MCC metrics values for the evaluation using the flDPnn training set and the CAID test set. The best values for each metric are boldfaced. The proposed DisorderUnetLM is underlined.

| Method | ROC-AUC | F1 | MCC |
| --- | --- | --- | --- |
| DisorderUnetLM | **0.826** | **0.516** | **0.414** |
| fIDPnn | 0.814 | 0.483 | 0.370 |
| fIDPlr | 0.793 | 0.452 | 0.330 |
| RawMSA | 0.780 | 0.445 | 0.328 |
| ESpritz-D | 0.774 | 0.428 | 0.307 |
| DisoMine | 0.765 | 0.424 | 0.299 |
| SPOT-Disorder2 | 0.760 | 0.469 | 0.349 |
| AUCpreD | 0.757 | 0.433 | 0.318 |
| SPOT-Disorder-Single | 0.757 | 0.432 | 0.315 |
| AUCpreD-np | 0.751 | 0.424 | 0.301 |
| Predisorder | 0.747 | 0.435 | 0.301 |

**Supplementary Table S4**. Detailed ROC-AUC values for the evaluation on the CheZOD test set. The best values are boldfaced. The proposed DisorderUnetLM is underlined.

| Method | ROC-AUC |
| --- | --- |
| DisorderUnetLM | **0.910** |
| SETH | **0.910** |
| ODiNPred | 0.890 |
| SPOT-Disorder-Single | 0.880 |
| AlphaFold2-rsa-25 | 0.880 |
| AUCpreD | 0.870 |
| MetaDisorder | 0.870 |
| MFDp2 | 0.850 |
| PrDOS | 0.840 |
| DISOPRED3 | 0.830 |
| flDPnn | 0.820 |

**Supplementary Table S5**. Detailed ROC-AUC values for the evaluation on the CAID-2 test sets, sorted by performance for the original Disorder-NOX subset. The best values for each subset are boldfaced. The proposed DisorderUnetLM is underlined. Missing values are for predictors with missing entries in the original CAID-2 predictions file.

| CAID-2 ROC-AUC | Original | | Non-redundant | |
|---|---|---|---|---|
| Method | NOX | PDB | NOX | PDB |
| <u>DisorderUnetLM</u> | <u>**0.844**</u> | <u>0.924</u> | <u>**0.834**</u> | <u>0.923</u> |
| Dispredict3 | 0.838 | 0.895 | 0.824 | 0.897 |
| fIDPnn2 | 0.836 | 0.894 | - | - |
| fIDPnn | 0.833 | 0.883 | 0.822 | 0.888 |
| fIDPtr | 0.827 | 0.899 | - | - |
| DisoPred | 0.821 | 0.919 | - | - |
| fIDPlr2 | 0.821 | 0.879 | - | - |
| IDP-Fusion | 0.818 | 0.933 | - | - |
| ESpritz-D | 0.802 | 0.899 | 0.788 | 0.898 |
| DeepIDP-2L | 0.800 | 0.922 | 0.786 | 0.920 |
| disomine | 0.797 | 0.886 | 0.782 | 0.888 |
| rawMSA | 0.783 | 0.910 | 0.714 | 0.859 |
| ENSHROUD-all | 0.780 | 0.906 | 0.767 | 0.906 |
| SPOT-Disorder2 | 0.780 | **0.949** | - | - |
| pyHCA | 0.779 | 0.898 | 0.771 | 0.909 |
| SETH-1 | 0.779 | 0.911 | - | 0.932 |
| SETH-0 | 0.772 | 0.930 | 0.768 | 0.912 |
| SPOT-Disorder | 0.772 | 0.931 | - | - |
| SPOT-Disorder-Single | 0.771 | 0.917 | 0.757 | 0.917 |
| AlphaFold-disorder | 0.767 | 0.929 | - | - |
| NetSurfP-3.0 | 0.766 | 0.933 | 0.751 | **0.933** |
| metapredict | 0.756 | 0.923 | 0.741 | 0.921 |
| IUPred3 | 0.755 | 0.885 | 0.743 | 0.887 |
| IsUnstruct | 0.748 | 0.885 | 0.736 | 0.886 |
| AlUPred | 0.747 | 0.903 | 0.756 | 0.906 |
| MobiDB-lite | 0.747 | 0.868 | 0.737 | 0.868 |
| ESpritz-X | 0.744 | 0.882 | 0.731 | 0.880 |
| VSL2 | 0.744 | 0.887 | 0.730 | 0.886 |
| PredIDR-long | 0.741 | 0.934 | 0.722 | 0.931 |
| PredIDR-short | 0.737 | 0.927 | 0.718 | 0.924 |
| ESpritz-N | 0.726 | 0.859 | 0.714 | 0.857 |
| PreDisorder | 0.726 | 0.893 | 0.766 | 0.899 |
| RONN | 0.723 | 0.857 | 0.666 | 0.769 |
| AUCpred-no-profile | 0.720 | 0.893 | 0.729 | 0.894 |
| AUCpred-profile | 0.706 | 0.922 | - | - |
| AlphaFold-rsa | 0.695 | 0.944 | - | - |
| DISOPRED3-diso | 0.692 | 0.879 | 0.679 | 0.875 |
| DisEMBL-dis465 | 0.680 | 0.792 | 0.673 | 0.792 |
| s2D | 0.680 | 0.771 | 0.759 | 0.930 |
| Dispredict2 | 0.644 | 0.671 | - | - |
| DisEMBL-disHL | 0.635 | 0.704 | 0.625 | 0.703 |
| FoldUnfold | 0.524 | 0.708 | 0.500 | 0.500 |
| APOD | 0.496 | 0.536 | 0.507 | 0.547 |